\documentclass[english]{article}
\usepackage[T1]{fontenc}
\usepackage[latin9]{inputenc}
\usepackage{geometry}
\usepackage{xcolor}
\usepackage{pdfcolmk}
\usepackage{bm}
\usepackage{amsmath}
\usepackage{amsthm}
\usepackage[authoryear]{natbib}
\PassOptionsToPackage{normalem}{ulem}
\usepackage{ulem}

\makeatletter

\providecolor{lyxadded}{rgb}{0,0,1}
\providecolor{lyxdeleted}{rgb}{1,0,0}

\theoremstyle{plain}
\newtheorem{thm}{\protect\theoremname}

\usepackage{nips15submit_e,times}
\usepackage{hyperref}
\usepackage{url}
\usepackage{braket}
\let\realket\Ket
\let\realbra\Bra

\nipsfinalcopy

\makeatother

\usepackage{babel}
\providecommand{\theoremname}{Theorem}

\begin{document}
\global\long\def\bb{\boldsymbol{B}}
\global\long\def\bt{\bm{\theta}}
\global\long\def\bmu{\bm{\mu}}
\global\long\def\bl{\bm{\lambda}}
\global\long\def\be{\bm{\eta}}
\global\long\def\E{\mathbb{E}}
\global\long\def\by{\bm{y}}
\global\long\def\bys{\bm{y^{\star}}}
\global\long\def\bSigma{\bm{\Sigma}}
\global\long\def\N{\mathcal{N}}
\global\long\def\I{\mathds{1}}
\global\long\def\A{\mathcal{A}}
\global\long\def\Ml{\mathcal{M}_{l}}
\global\long\def\Q{\mathcal{Q}}
\global\long\def\epinf{EP^{\infty}}
\global\long\def\O{\mathcal{O}}
\global\long\def\G{\mathcal{G}}
\global\long\def\bS{\boldsymbol{S}}
\global\long\def\Bra#1{\realbra{#1}}
\global\long\def\Ket#1{\realket{#1}}

\title{Expectation Propagation performs a smoothed gradient descent}

\author{Guillaume Dehaene\\
Ecole Polytechnique Federale de Lausanne\\
\texttt{guillaume.dehaene@gmail.com}}
\maketitle
\begin{abstract}
Bayesian inference is a popular method to build learning algorithms
but it is hampered by the fact that its key object, the posterior
probability distribution, is often uncomputable. Expectation Propagation
(EP) (\citet{Minka:EP}) is a popular algorithm that solves this issue
by computing a parametric approximation (e.g: Gaussian) to the density
of the posterior. However, while it is known empirically to quickly
compute fine approximations, EP is extremely poorly understood which
prevents it from being adopted by a larger fraction of the community.

The object of the present article is to shed intuitive light on EP,
by relating it to other better understood methods. More precisely,
we link it to using gradient descent to compute the Laplace approximation
of a target probability distribution. We show that EP is exactly equivalent
to performing gradient descent on a smoothed energy landscape: i.e:
the original energy landscape convoluted with some smoothing kernel.
This also relates EP to algorithms that compute the Gaussian approximation
which minimizes the reverse KL divergence to the target distribution,
a link that has been conjectured before but has not been proved rigorously
yet. These results can help practitioners to get a better feel for
how EP works, as well as lead to other new results on this important
method.

This article was submitted and accepted to the Advances in Approximate
Bayesian Inference NIPS 2016 workshop (www.approximateinference.org).
\end{abstract}
Throughout this article, we consider the task of approximating a probability
density over a one-dimensional space:
\begin{align}
p\left(\theta\right) & =\exp\left(-\psi\left(\theta\right)\right)
\end{align}
where we will assume for simplicity that $\psi\left(\theta\right)$
is convex so that $\psi^{''}\left(\theta\right)>0$.

We will present various algorithms to compute Gaussian approximations
of $p$. We will first seek to compute the ``Laplace approximation''.
We then turn to computing the Gaussian which minimizes the reverse
KL divergence to $p\left(\theta\right)$. Finally, we consider using
Expectation Propagation (EP). These three approximations can be computed
by similar algorithms: either by exact gradient descent or by variants
to it which we refer to as ``smoothed gradient descent''. This new
perspective on EP is useful in giving practitioners a more intuitive
framework than current justifications of EP, but it can also be used
to derive new theoretical results on the method. For example, we show
that, in various asymptotes, the limit behavior of EP is simple.

To keep the presentation of these ideas sharp and light, we do not
prove the results we present in this document. All proofs, which are
stated in the high-dimensional case, can be found in the appendix.

\section{The Laplace approximation}

If we want to compute a Gaussian approximation of $p\left(\theta\right)$,
a natural idea consists in performing a Taylor expansion to second
order of $\psi\left(\theta\right)$. It seems natural to perform the
expansion around the global maximum $\theta^{\star}$ of the target
$p\left(\theta\right)$ (which is unique because of the convexity
of $\psi\left(\theta\right)$). This yields the ``Laplace'' approximation
of the target (\citet{Murphy:2012:MLP:2380985} 8.4.1):
\begin{equation}
p\left(\theta\right)\approx p\left(\theta^{\star}\right)\exp\left(-\psi^{''}\left(\theta^{\star}\right)\frac{\left(\theta^{\star}-\theta\right)^{2}}{2}\right)
\end{equation}

This requires us to compute the global maximum $\theta^{\star}$ (which
is unique under our convexity assumption). A nice solution for this
is Newton's method (NT), which corresponds to gradient descent of
$\psi\left(\theta\right)$ with a Hessian correction. This method
is actually extremely closely linked to the ``Laplace'' approximation,
since we can derive it as iterating over Gaussian approximations.
Indeed, consider the following method:
\begin{itemize}
\item Start from some initial Gaussian approximation $q^{\left(0\right)}\left(\theta\right)$
to $p\left(\theta\right)$, with mean $\mu_{0}$.
\item Then loop until convergence:
\begin{enumerate}
\item Compute the mean $\mu^{\left(t\right)}$ of the current approximation
$q^{\left(t\right)}\left(\theta\right)$.
\item Construct the new approximation using the second degree expansion
of $\psi$ around $\mu^{\left(t\right)}$:
\begin{equation}
q^{\left(t+1\right)}\left(\theta\right)\propto\exp\left(-\psi^{'}\left(\mu^{\left(t\right)}\right)\left(\theta-\mu^{\left(t\right)}\right)-\psi^{''}\left(\mu^{\left(t\right)}\right)\frac{\left(\theta-\mu^{\left(t\right)}\right)^{2}}{2}\right)\label{eq: the new Gaussian approximation}
\end{equation}
\end{enumerate}
\end{itemize}
It is straightforward to check that the dynamics of $\mu^{\left(t\right)}$
in this algorithm match those in the classical NT algorithm. The fixed-point
of the Gaussian-iterating algorithm is the Laplace approximations
of the target $p\left(\theta\right)$.

Gradient descent is fairly easy to understand from an intuitive point
of view. It corresponds roughly to the dynamics of a ball which we
drop on some complicated energy landscape $\psi\left(\theta\right)$
and falls down along the valleys of the landscape until it reaches
some local minimum.

In practice, the Laplace approximation does not provide a very good
approximation of the target $p\left(\theta\right)$. Intuitively,
this is because it depends on the value of $\psi$ at a single point
$\theta^{\star}$ so it naturally fails at providing a global account
of the target distribution.

\section{Smoothing the gradient minimizes the reverse KL divergence}

In order to provide a Gaussian approximation that gives a more global
fit to the target distribution, why not replace the point estimates
$\psi^{'}\left(\mu^{\left(t\right)}\right)$ and $\psi^{''}\left(\mu^{\left(t\right)}\right)$
by averages over a large region of $\theta$-space? We could use many
densities for this, but it seems natural to use the current Gaussian
approximation $q^{\left(t\right)}\left(\theta\right)$ which is indeed
centered at $\mu^{\left(t\right)}$.

We can thus construct a ``smoothed Newton method'' by replacing
eq. \eqref{eq: the new Gaussian approximation} with:
\begin{equation}
q^{\left(t+1\right)}\left(\tilde{\theta}\right)\propto\exp\left(-E_{q^{\left(t\right)}}\left[\psi^{'}\left(\theta\right)\right]\left(\tilde{\theta}-\mu^{\left(t\right)}\right)-E_{q^{\left(t\right)}}\left[\psi^{''}\left(\theta\right)\right]\frac{\left(\tilde{\theta}-\mu^{\left(t\right)}\right)^{2}}{2}\right)\label{eq: the VB Gaussian iteration}
\end{equation}

This iteration has further interesting properties. First of all, the
fixed-point of this iteration is unique under our assumption that
$\psi\left(\theta\right)$ is convex (\citet{challis2011concave}).
Furthermore, even when $\psi\left(\theta\right)$ is not convex, the
ensemble of all fixed-points of the iteration is also the ensemble
of extrema of the reverse Kullback-Leibler divergence on the space
of Gaussians:
\begin{equation}
KL\left(q,p\right)=E_{q}\left(\log\frac{q\left(\theta\right)}{p\left(\theta\right)}\right)\label{eq: the reverse KL divergence}
\end{equation}

In other words, the iterative algorithm we have defined computes a
Gaussian Variational Bayes (VB) approximation of $p\left(\theta\right)$
(\citet{hoffman2013stochastic}). This link between gradient descent
and Variational Bayes was originally derived by \citet{opper2009variational}.

Critically for the rest of this article, the computation of the expected
value of the second derivative can be rewritten:
\begin{equation}
E_{q^{\left(t\right)}}\left[\psi^{''}\left(\theta\right)\right]=\left[\text{var}_{q^{\left(t\right)}}\right]^{-1}E_{q^{\left(t\right)}}\left[\psi^{'}\left(\theta\right)\left(\theta-\mu^{\left(t\right)}\right)\right]
\end{equation}
This equality is found by integration by parts and is only true for
a Gaussian distribution. When we consider non-Gaussian kernels in
the following sections, we will adapt the updating equation \eqref{eq: the VB Gaussian iteration}
using this second form for the quadratic term.

\section{Hybrid smoothing minimizes the alpha-divergence}

It might seem weird to use a Gaussian smoothing: couldn't we use something
that is closer to the target distribution? We will do so by geometrically
mixing the target and the current Gaussian approximation, thus building
a ``hybrid'' distribution. For some $0<\alpha<1$, construct the
$\alpha$-hybrid (with $Z_{\alpha}^{\left(t\right)}$ the normalizing
constant):
\begin{equation}
h_{\alpha}^{\left(t\right)}\left(\theta\right)=\left(Z_{\alpha}^{\left(t\right)}\right)^{-1}\left(q^{\left(t\right)}\left(\theta\right)\right)^{\alpha}\left(p\left(\theta\right)\right)^{1-\alpha}\label{eq: the alpha-divergence hybrid}
\end{equation}
which we then use as the smoothing kernel instead of $q^{\left(t\right)}\left(\theta\right)$.
We also change the centering point for the approximation: the VB update,
eq. \eqref{eq: the VB Gaussian iteration}, uses the mean of the current
Gaussian approximation; in our new update, we will use the mean of
the hybrid distribution instead. The update corresponding to the $\alpha$-hybrid
is then, noting $\mu_{h}=E_{h_{\alpha}^{\left(t\right)}}\left[\theta\right]$
and $v_{h}=\text{var}_{h_{\alpha}^{\left(t\right)}}$:

\begin{align}
q^{\left(t+1\right)}\left(\tilde{\theta}\right)\propto & \exp\left(-E_{h_{\alpha}^{\left(t\right)}}\left[\psi^{'}\left(\theta\right)\right]\left(\tilde{\theta}-\mu_{h}\right)-v_{h}^{-1}E_{h_{\alpha}^{\left(t\right)}}\left[\psi^{'}\left(\theta\right)\left(\theta-\mu_{h}\right)\right]\frac{\left(\tilde{\theta}-\mu_{h}\right)^{2}}{2}\right)\label{eq: the alpha hybrid-iteration}
\end{align}

Once again, this corresponds to both a smoothed Newton's method (by
construction) but also, much more surprisingly, to an algorithm that
minimizes a specific divergence. Indeed, all fixed-points of the $\alpha$-hybrid-smoothing
iteration eq. \eqref{eq: the alpha hybrid-iteration} are also extrema
of the $\alpha$-divergence (which represents a smooth interpolation
between the reverse KL-divergence at $\alpha=1$ and the forward KL-divergence
$KL\left(p,q\right)$ at $\alpha=0$; \citet{Minka:DivMeasuresMP}).

\section{Classical Expectation Propagation is a smoothed gradient method}

In this article, we have shown that we can find extrema of all $\alpha$-divergences
and, critically, of the reverse KL-divergence by performing a form
of smoothed gradient descent (with a Hessian correction). This might
shed some light on these methods and help further theoretical investigation
of these methods. However, the main interest of this approach consists
in using this idea of smoothed gradient to give a justification of
Expectation Propagation (EP, \citet{Minka:EP}) which is much more
intuitively satisfying than current derivations of this method (which
we will assume that the reader is familiar with due to space constraints;
see \citet{Minka:EP,Seeger:EPExpFam,BishopPRML}).

In order to apply EP, we have to assume that the target distribution
factorizes into ``simple'' factor functions $f_{i}\left(\theta\right)$:
\begin{equation}
p\left(\theta\right)=\prod_{i=1}^{n}f_{i}\left(\theta\right)
\end{equation}

We will note $\phi_{i}\left(\theta\right)=-\log\left(f_{i}\left(\theta\right)\right)$,
thus the energy landscape for the gradient descent $\psi\left(\theta\right)$
has been split into $n$ additive components:
\begin{equation}
\psi\left(\theta\right)=\sum_{i}\phi_{i}\left(\theta\right)\label{eq: splitting the energy landscape}
\end{equation}

All algorithms we have presented so far compute a single Gaussian
approximation for the whole target distribution. This next algorithm
will compute for each time-step a ``local'' Gaussian approximation
for each single factor $q_{i}^{\left(t\right)}\approx f_{i}$. We
can then compute a global approximation of the target by combining
them multiplicatively:
\begin{equation}
q_{\text{global}}^{\left(t\right)}=\prod_{i=1}^{n}q_{i}^{\left(t\right)}\approx p\label{eq: the global approximation}
\end{equation}

Each local Gaussian approximations $g_{i}^{\left(t\right)}$ is updated
by a smoothed gradient descent on the corresponding local energy landscape
$\phi_{i}$. For the smoothing, we construct once more a hybrid distribution.
The hybrid for the update of the $i^{th}$ approximation is formed
by multiplying the true factor $f_{i}$ and the current local Gaussian
approximation of all other factors $\left(g_{i}^{\left(t\right)}\right)_{j\neq i}$.
I.e:
\begin{equation}
h_{i}^{\left(t\right)}\left(\theta\right)=\left(Z_{i}^{\left(t\right)}\right)^{-1}f_{i}\left(\theta\right)\prod_{j\neq i}g_{j}^{\left(t\right)}\left(\theta\right)\label{eq: defining the EP-hybrid}
\end{equation}

The new Gaussian approximation of the factor $f_{i}$ is then given
by, noting $\mu_{h_{i}}=E_{h_{i}^{\left(t\right)}}\left[\theta_{h}\right]$
and $v_{h_{i}}=\text{var}_{h_{i}^{\left(t\right)}}$:

\begin{align}
g_{i}^{\left(t+1\right)}\left(\theta\right)\propto & \exp\left(-E_{h_{i}^{\left(t\right)}}\left[\phi_{i}^{'}\left(\theta_{h}\right)\right]\left(\theta-\mu_{h_{i}}\right)-v_{h_{i}}^{-1}E_{h_{i}^{\left(t\right)}}\left[\phi_{i}^{'}\left(\theta_{h}\right)\left(\theta_{h}-\mu_{h_{i}}\right)\right]\frac{\left(\theta-\mu_{h_{i}}\right)^{2}}{2}\right)\label{eq: the local update in EP}
\end{align}

\textbf{\uline{This iterating scheme is exactly the same as the
classical EP update}} presented by \citet{Minka:EP}. We have thus
rephrased the EP iteration from its original computationally-convenient
but cryptic form into a smoothed gradient descent which is much more
instructive for our intuitive understanding of EP. 

\section{Why this matters}

\subsection{Asymptotic behavior}

This new perspective on VB, $\alpha$-divergence minimization and
EP can be used to derive several interesting results on the asymptotic
behavior of these methods in a painless manner. Indeed, as we have
shown, these algorithms all correspond to a smoothed gradient descent.
Thus, in all limits in which the smoothing kernel is sufficiently
concentrated, the dynamics of these algorithms asymptote to the dynamics
of Newton's method.

Furthermore, in all limits in which the smoothing kernel of one method
asymptotes to the smoothing kernel of a second, their dynamics asymptote
to one another. A trivial example of this is the fact that the dynamics
of $\alpha$-hybrid smoothing (eq. \eqref{eq: the alpha-divergence hybrid})
asymptote to the dynamics of the Gaussian smoothing (eq. \eqref{eq: the VB Gaussian iteration})
in the limit $\alpha\rightarrow1$ (thus giving a much more intuitive
understanding of the result of \citet{dehaene2015expectation}). A
much more interesting example consists in proving a ``folk theorem''
on EP asserting that, in the limit of a large number of sites with
each one having a negligible contribution to the ensemble, EP corresponds
to minimizing the reverse KL divergence (eq. \eqref{eq: the reverse KL divergence}).
The present work shows painlessly that this is true in any limit in
which all EP hybrids $h_{i}$ (eq. \eqref{eq: defining the EP-hybrid})
asymptote to the current global approximation $q_{\text{global}}=\frac{q_{i}}{f_{i}}h_{i}$
(eq. \eqref{eq: the global approximation}).

\subsection{Intuitive understanding}

In this article, we have unified several algorithms as performing
smoothed gradient descent. While this might provide the path towards
more useful algorithms for minimizing several divergence measures,
we believe that the most useful contribution of this work is that
it gives practitioners a more intuitive understanding of these algorithms.
Indeed, gradient descent is both a fairly intuitive algorithm, matching
our physical intuitions of an object sliding around on some energy
landscape, and one which has been extremely extensively researched.
The addition of smoothing complicates this picture, but only slightly.
We thus hope that our results can help promote these methods by shining
a new light on them. This contribution is probably the most helpful
for EP, given that our iteration (eq. \eqref{eq: the local update in EP})
and classical EP (\citet{Minka:EP}) \textbf{\uline{have the exact
same dynamics}}, and that current presentations of EP give no intuitions
about how the algorithm operates.

\bibliographystyle{plainnat}
\bibliography{refNipsWorkshop2016}

\appendix
\newpage{}

\section*{Appendix}

In this appendix, we will present all of the technical results underlying
our main text, which was light on details so as to remain compact.

Throughout this document, we will present various methods to compute
a Gaussian approximation of a multivariate target distribution:
\begin{equation}
p\left(\bt\right)=\exp\left(-\psi\left(\bt\right)\right)
\end{equation}

We will note $d$ the dimensionality of the space. All vectors will
be represented by bold letters (e.g: $\bt,\bmu$) while matrices are
bold capitalized letters (e.g: $\bS,\bb$).

In contexts where it matters to distinguish the inner and outer-product
between vectors, we use the physicists Bra-Ket notation which makes
it obvious whether a given vector should be considered as a $\left(1*d\right)$
matrix (in which it is called a ``Bra'' $\Bra{\bt}$) or a $\left(d*1\right)$
matrix (in which case it is called a ``Ket'' $\Ket{\bmu})$. In
this notation, scalar products are represented as, for example: $\Bra{\bt}\Ket{\bmu}=\sum_{i=1}^{d}\theta_{i}\mu_{i}$.
A scalar product weighted by some matrix $\bb$ is represented using:
$\Bra{\bt}\bb\Ket{\bmu}=\sum_{i,j}\theta_{i}\mu_{j}B_{i,j}$. Outer-products
are noted as $\Ket{\bt}\Bra{\bmu}$ (which corresponds to the conventional
notation $\bt\bmu^{T}$) which gives a matrix: $\left(\Ket{\bt}\Bra{\bmu}\right)_{i,j}=\theta_{i}\mu_{j}$.

Finally, for a multivariate function such as $p\left(\bt\right)$,
we note $\nabla_{\bt}p\left(\bt\right)$ the gradient, i.e: the vector
of the derivatives against each component of the variable. We note
$\boldsymbol{H}p\left(\bt\right)$ the ``Hessian'' matrix of the
second derivatives of the function.

\section{The relationship between Variational Bayes and smoothed gradient
descent}

In this first section, we show that we can minimize the reverse KL
divergence between a Gaussian approximation and the target distribution
by performing smoothed gradient descent.

\subsection{The Gaussian Variational Bayes approximation}

First, let us consider computing a Gaussian approximation which minimizes
the ``reverse'' KL divergence to the target. Noting $\G$ the space
of all Gaussians, this Gaussian Variational Bayes approximation is
defined as:
\begin{align}
q_{\text{VB}}\left(\bt\right) & =\text{argmin}_{q\in\mathcal{\G}}\left(q,p\right)\\
 & =\text{argmin}_{q\in\mathcal{\G}}\int q(\bt)\log\left[\frac{q\left(\bt\right)}{p\left(\bt\right)}\right]\label{eq: VB objective function}
\end{align}

Let us now rewrite the objective function. The most common parameterization
of a Gaussian distribution is via its mean and its covariance matrix.
However, an alternative parameterization which is more relevant in
this case is using the ``matrix square root'' of the covariance
(which, in 1D, would correspond to the standard-deviation). Noting
$\bSigma$ the covariance matrix, this ``matrix square root'' is
a solution of:
\begin{equation}
\bS\bS^{T}=\bSigma
\end{equation}

This parameterization is useful as it enables us to write any Gaussian
$\boldsymbol{g}_{\bmu,\boldsymbol{S}}$with parameters $\left(\bmu,\bS\right)$
as a translated and shifted version of a Gaussian with mean 0 and
covariance the identity matrix (which we note $\boldsymbol{g}_{0}$):
\begin{equation}
\boldsymbol{g}_{\bmu,\boldsymbol{S}}=\boldsymbol{S}\boldsymbol{g}_{0}+\bmu
\end{equation}
However, note that this parameterization of the space of Gaussians
is degenerate: the same Gaussian density corresponds to multiple values
of $\bS$.

Using this parameterization, it is easy to write down the reverse
KL divergence as an expected value under the random variable $\boldsymbol{g}_{0}$:
\begin{align}
KL\left(q_{\bmu,\boldsymbol{S}},p\right) & =\int q(\bt)\log\left[\frac{q\left(\bt\right)}{p\left(\bt\right)}\right]\\
 & =\int q(\bt)\psi\left(\bt\right)+\int q\left(\bt\right)\log\left(q\left(\bt\right)\right)\\
 & =E\left[\psi\left(\boldsymbol{S}\boldsymbol{g}_{0}+\bmu\right)\right]-\log\left(\left(2\pi\right)^{d/2}\left|\boldsymbol{S}\right|\right)\label{eq: rewritting the KL divergence as a trivially convex objective}
\end{align}
where $\left|\boldsymbol{S}\right|$ is the determinant of the $\boldsymbol{S}$
matrix.

Computing the gradient of the KL divergence against this parameterization
of the Gaussians is then a straightforward exercise in vectorial and
matrix derivatives (cf: the matrix cookbook \citet{IMM2012-03274})
which yields:
\begin{align}
\nabla_{\bmu}KL\left(q_{\bmu,\boldsymbol{S}},p\right) & =E\left[\nabla\psi\left(\boldsymbol{S}\boldsymbol{g}_{0}+\bmu\right)\right]\\
 & =E\left[\nabla\psi\left(\boldsymbol{g}_{\bmu,\boldsymbol{S}}\right)\right]\\
\nabla_{\boldsymbol{S}}KL\left(q_{\bmu,\boldsymbol{S}},p\right) & =E\left[\Ket{\nabla\psi\left(\bS\boldsymbol{g}_{0}+\bmu\right)}\Bra{\boldsymbol{g}_{0}}\right]-\bS^{-1}\label{eq: we will IPP this equation}
\end{align}
We then integrate by parts eq. \eqref{eq: we will IPP this equation}
(or, alternatively, we use Stein's lemma) to obtain an equation with
the second derivative of $\psi$:
\begin{align}
\nabla_{\boldsymbol{S}}KL\left(q_{\bmu,\boldsymbol{S}},p\right) & =E\left[\boldsymbol{H}\psi\left(\bS\boldsymbol{g}_{0}+\bmu\right)\boldsymbol{S}^{T}\right]-\bS^{-1}\\
 & =E\left[\boldsymbol{H}\psi\left(\boldsymbol{g}_{\bmu,\boldsymbol{S}}\right)\right]\bS^{T}-\bS^{-1}
\end{align}

This gives us a simple characterization of all critical points of
the function $\bmu,\bS\rightarrow KL$: they obey the following equations:
\begin{align}
E\left[\nabla\psi\left(\boldsymbol{g}_{\bmu,\boldsymbol{S}}\right)\right] & =\overrightarrow{0}\label{eq: smoothed gradient is 0}\\
E\left[\boldsymbol{H}\psi\left(\boldsymbol{g}_{\bmu,\boldsymbol{S}}\right)\right] & =\bS^{-1}\left(\bS^{T}\right)^{-1}\\
 & =\boldsymbol{\Sigma}^{-1}\label{eq: smoothed second derviative gives precision}
\end{align}

These conditions for the Gaussian Variational Bayes approximation
of a target distribution were originally derived by \citet*{opper2009variational}.
They express that a Gaussian VB approximation must be such that the
expected value of the log-gradient of the target distribution is 0
in all dimensions and that the inverse-variance of the approximation
matches the expected value of the log-Hessian of the target distribution.

These conditions are quite naturally linked to the smoothed gradient
descent algorithm which we now introduce.

\subsection{Smoothed gradient descent}

First, let us present the smoothed gradient descent (with a Hessian
correction) in detail. It corresponds to the following algorithm.
\begin{itemize}
\item Initialize the algorithm with any Gaussian approximation $q^{\left(0\right)}$
of the target distribution
\item Then loop until convergence:
\begin{enumerate}
\item Compute the mean as well as the expected log-gradient and expected
log-Hessian of the target distribution under the current Gaussian
approximation $q^{\left(t\right)}\left(\bt\right)$:
\begin{align}
\bmu & =E_{q^{\left(t\right)}}\left[\bt\right]\\
E\nabla & =E_{q^{\left(t\right)}}\left[\nabla\psi\left(\bt\right)\right]\\
E\boldsymbol{H} & =E_{q^{\left(t\right)}}\left[\boldsymbol{H}\psi\left(\bt\right)\right]
\end{align}
\item Compute the new Gaussian approximation using the following formula:
\begin{equation}
q^{\left(t+1\right)}=\frac{1}{Z}\exp\left(-\Bra{E\nabla}\Ket{\left(\bt-\bmu\right)}-\Bra{\left(\bt-\bmu\right)}\frac{E\boldsymbol{H}}{2}\Ket{\left(\bt-\bmu\right)}\right)\label{eq: the GVB updating equation}
\end{equation}
where $Z$ is a normalizing constant.
\end{enumerate}
\end{itemize}
We refer to this algorithm as ``smoothed gradient descent'' since
this update exactly matches the update of gradient descent with a
Hessian correction starting from $\bmu=E_{q^{\left(t\right)}}\left[\bt\right]$
on the energy landscape given by a smoothing of $\psi$ with a Gaussian
kernel $q^{\left(t\right)}$ (in equations: $\bt\rightarrow\left(\psi\otimes q^{\left(t\right)}\right)\left(\bt\right)$
is the new energy landscape).

This algorithm does not correspond to performing gradient descent
on the space Gaussians parameterized by $\left(\bmu,\bS\right)$ of
the reverse KL divergence, but it is closely related as these two
algorithms share their fixed-points. Indeed, considering eq. \eqref{eq: the GVB updating equation}
shows that this algorithm is stable if and only if:
\begin{itemize}
\item $E\nabla=0$ so that the mean of the Gaussian approximation does not
change.
\item $E\boldsymbol{H}=\left(\text{Cov}_{q}\right)^{-1}$ so that the variance
of the distribution does not change either.
\end{itemize}
These stability conditions exactly match the characterization of critical
points of $\bmu,\bS\rightarrow KL$ (eqs. \eqref{eq: smoothed gradient is 0}
and \eqref{eq: smoothed second derviative gives precision}). Thus,
this smoothed gradient descent algorithm represents an iterative scheme
to minimize the reverse KL divergence over the space of Gaussian distributions.

\section{Minimizing the $\alpha$-divergence}

In this second section, we introduce the $\alpha$-divergence, which
interpolates between the forward KL divergence and the reverse KL
divergence. We show that we can minimize the $\alpha$-divergence
between a Gaussian approximation and the target distribution by performing
a smoothed gradient descent which uses an hybrid distribution as the
smoothing kernel.

\subsection{Defining the $\alpha$-divergence}

Consider two probability densities $\left(p\left(\bt\right),q\left(\bt\right)\right)$.
The $\alpha$-divergence between these two distributions is given,
for $\alpha\notin\left\{ 0,1\right\} $:
\begin{equation}
D_{\alpha}\left(p,q\right)=\frac{1}{\alpha\left(1-\alpha\right)}\left[1-\int\left(p\left(\bt\right)\right)^{1-\alpha}\left(q\left(\bt\right)\right)^{\alpha}d\bt\right]\label{eq: defining the alpha divergence}
\end{equation}

A few values of $\alpha$ correspond to interesting measures of the
differences between $p$ and $q$ (\citet{Minka:DivMeasuresMP}; note
that we are not using the exact same definition of the $\alpha$-divergence).
\begin{itemize}
\item For $\alpha=1/2$, we recover the squared Hellinger distance:
\begin{align}
D_{1/2} & =4\left[1-\int\sqrt{pq}\right]\\
 & =2\int\left(\sqrt{p}-\sqrt{q}\right)^{2}
\end{align}
\item For $\alpha=-1$ and $\alpha=2$, we recover interesting values:
\begin{align}
D_{-1} & =-\frac{1}{2}\left[1-\int\frac{p^{2}}{q}\right]\\
 & =\frac{1}{2}\left[\int\frac{p^{2}}{q}+\int q-2\int p\right]\\
 & =\frac{1}{2}\int\frac{\left(p-q\right)^{2}}{q}\\
D_{2} & =\frac{1}{2}\int\frac{\left(p-q\right)^{2}}{p}
\end{align}
This corresponds to the two $\chi^{2}$ distances between the two
probability distributions
\item Finally, and most interesting, in the limits $\alpha\rightarrow0$
and $\alpha\rightarrow1$, we recover the direct and reverse KL divergences:
\begin{align}
D_{\alpha}\left(p,q\right) & =\frac{1}{\alpha\left(1-\alpha\right)}\left[1-\int p\left(\frac{p}{q}\right)^{-\alpha}\right]\\
 & =\frac{1}{\alpha\left(1-\alpha\right)}\left[1-\int p\exp\left(-\alpha\log\frac{p}{q}\right)\right]\\
 & \approx\frac{1}{\alpha\left(1-\alpha\right)}\left[1-\int p\left(1+-\alpha\log\frac{p}{q}+\alpha^{2}\dots\right)\right]\\
 & \approx\frac{1}{1-\alpha}\int p\log\frac{p}{q}+\alpha\dots\\
\lim_{\alpha\rightarrow0}D_{\alpha}\left(p,q\right) & =\int p\log\frac{p}{q}\\
 & =KL\left(p,q\right)\\
\lim_{\alpha\rightarrow1}D_{\alpha}\left(p,q\right) & =KL\left(q,p\right)
\end{align}
This is the result we are most interested in, since it justifies our
earlier remark that the $\alpha$-divergences interpolate between
the forward and reverse KL divergences.
\end{itemize}

\subsection{Minimizing the $\alpha$-divergence over the space of Gaussians}

Once more, we will parametrize Gaussians using the mean and the square-root
matrix of the covariance $\bS$. We can then smartly rewrite our objective
function, the $\alpha$-divergence, by performing a change of variable
$\tilde{\bt}=\bS^{-1}\left(\bt-\bmu\right)$. In this reference frame,
the Gaussian density is constant. This gives:
\begin{equation}
D_{\alpha}\left(p,q_{\bmu,\boldsymbol{S}}\right)=\frac{1}{\alpha\left(1-\alpha\right)}\left[1-\int p\left(\bS\tilde{\bt}+\bmu\right)^{1-\alpha}q_{0}\left(\tilde{\bt}\right)^{\alpha}\frac{d\tilde{\bt}}{\left|\bS\right|^{1-\alpha}}\right]
\end{equation}

We can then compute the gradients:
\begin{equation}
\nabla_{\bmu}D_{\alpha}=\frac{1}{\alpha\left(1-\alpha\right)}\int\left(1-\alpha\right)\nabla\psi\left(\bS\tilde{\bt}+\bmu\right)p\left(\bS\tilde{\bt}+\bmu\right)^{1-\alpha}q_{0}\left(\tilde{\bt}\right)^{\alpha}\frac{d\tilde{\bt}}{\left|\bS\right|^{1-\alpha}}
\end{equation}

By now performing the reverse change of variable and returning to
$\bt$, we get that the gradient corresponds to an expected value
under the hybrid distribution: $h_{\alpha}\left(\bt\right)=Z_{\alpha}^{-1}\left(p\left(\bt\right)\right)^{1-\alpha}\left(q_{\bmu,\bS}\left(\bt\right)\right)^{\alpha}$:
\begin{align}
\nabla_{\bmu}D_{\alpha} & =\frac{1}{\alpha}\int\nabla\psi\left(\bt\right)p\left(\bt\right)^{1-\alpha}q_{\bmu,\bS}\left(\bt\right)^{\alpha}d\bt\\
 & =\frac{\left(\int\left(p\left(\bt\right)\right)^{1-\alpha}\left(q_{\bmu,\bS}\left(\bt\right)\right)^{\alpha}d\bt\right)}{\alpha}E_{h_{\alpha}}\left(\nabla\psi\left(\bt\right)\right)
\end{align}

Similarly, the gradient to $\bS$ gives another expected value under
$h_{\alpha}$:
\begin{align}
\nabla_{\bS}D_{\alpha}= & \frac{1}{\alpha}\int\Ket{\tilde{\bt}}\Bra{\nabla\psi\left(\bS\tilde{\bt}+\bmu\right)}p\left(\bS\tilde{\bt}+\bmu\right)^{1-\alpha}q_{0}\left(\tilde{\bt}\right)^{\alpha}\frac{d\tilde{\bt}}{\left|\bS\right|^{1-\alpha}}\nonumber \\
 & -\frac{1}{\alpha}\bS^{-1}\frac{1}{\left|\bS\right|^{1-\alpha}}\int p\left(\bS\tilde{\bt}+\bmu\right)^{1-\alpha}q_{0}\left(\tilde{\bt}\right)^{\alpha}d\tilde{\bt}\\
= & \frac{1}{\alpha}\bS^{-1}\int\Ket{\bt-\bmu}\Bra{\nabla\psi\left(\bt\right)}p\left(\bt\right)^{1-\alpha}q_{\bmu,\bS}\left(\bt\right)^{\alpha}d\bt\nonumber \\
 & -\frac{1}{\alpha}\bS^{-1}\left(\int\left(p\left(\bt\right)\right)^{1-\alpha}\left(q_{\bmu,\bS}\left(\bt\right)\right)^{\alpha}d\bt\right)\\
= & \frac{\left(\int\left(p\left(\bt\right)\right)^{1-\alpha}\left(q_{\bmu,\bS}\left(\bt\right)\right)^{\alpha}d\bt\right)}{\alpha}\bS^{-1}\left[E_{h_{\alpha}}\left(\Ket{\bt-\bmu}\Bra{\nabla\psi\left(\bt\right)}\right)-\boldsymbol{I}_{d*d}\right]
\end{align}

Thus the critical points of the function $\bmu,\bS\rightarrow D_{\alpha}\left(p,q_{\bmu,\bS}\right)$
obey the following two equations:
\begin{align}
E_{h_{\alpha}}\left(\nabla\psi\left(\bt\right)\right) & =0\label{eq: smoothed gradient under the hybrid is 0}\\
E_{h_{\alpha}}\left(\Ket{\bt-\bmu}\Bra{\nabla\psi\left(\bt\right)}\right) & =\boldsymbol{I}_{d*d}\label{eq: smoothed the other thing is identity}
\end{align}

Furthermore, for any probability distribution (with fast decrease
in the tails) by integration by parts we have that (illustrating the
result with $p\left(\bt\right)\propto\exp\left(-\psi\left(\bt\right)\right)$)
$\forall\boldsymbol{v}$:
\begin{align}
E_{p}\left(\nabla\psi\left(\bt\right)\right) & =0\\
E_{p}\left(\Ket{\bt-\boldsymbol{v}}\Bra{\nabla\psi\left(\bt\right)}\right) & =\boldsymbol{I}_{d*d}\label{eq: the second Stein equation}
\end{align}

By applying these two equations to the hybrid distribution at a critical
point $h_{\star}=p^{1-\alpha}\left(q_{\bmu_{\star},\bS_{\star}}\right)^{\alpha}$
with the two eqs. \eqref{eq: smoothed gradient under the hybrid is 0}
and \eqref{eq: smoothed the other thing is identity}, we get that
at a critical point, the mean and variance of the hybrid and of the
Gaussian approximation are identical. Noting $\bSigma_{\star}=\bS_{\star}\left(\bS_{\star}\right)^{T}$
the covariance of the Gaussian distribution at the critical point:
\begin{align}
0 & =E_{h_{\star}}\left(\left(1-\alpha\right)\nabla\psi\left(\bt\right)+\alpha\bSigma_{\star}^{-1}\left(\bt-\bmu_{\star}\right)\right)\\
 & =0+E_{h_{\star}}\left(\alpha\bSigma_{\star}^{-1}\left(\bt-\bmu_{\star}\right)\right)\\
 & =\alpha\bSigma_{\star}^{-1}\left(E_{h_{\star}}\left(\bt\right)-\bmu_{\star}\right)\\
E_{h_{\star}}\left(\bt\right) & =\bmu_{\star}
\end{align}

And (using $\boldsymbol{v}=\bmu_{\star}$ in eq. \ref{eq: the second Stein equation}):
\begin{align}
\boldsymbol{I}_{d*d} & =E_{h_{\star}}\left(\Ket{\bt-\bmu_{\star}}\Bra{\left(\left(1-\alpha\right)\nabla\psi\left(\bt\right)+\alpha\bSigma_{\star}^{-1}\left(\bt-\bmu_{\star}\right)\right)}\right)\\
 & =\left(1-\alpha\right)E_{h_{\star}}\left(\Ket{\bt-\bmu_{\star}}\Bra{\nabla\psi\left(\bt\right)}\right)+\alpha E_{h_{\star}}\left(\Ket{\bt-\bmu_{\star}}\Bra{\bt-\bmu_{\star}}\bSigma_{\star}^{-1}\right)\\
 & =\left(1-\alpha\right)\boldsymbol{I}_{d*d}+\alpha E_{h_{\star}}\left(\Ket{\bt-\bmu_{\star}}\Bra{\bt-\bmu_{\star}}\bSigma_{\star}^{-1}\right)\\
\alpha\boldsymbol{I}_{d*d} & =\alpha E_{h_{\star}}\left(\Ket{\bt-\bmu_{\star}}\Bra{\bt-\bmu_{\star}}\bSigma_{\star}^{-1}\right)\\
\boldsymbol{I}_{d*d} & =\text{\textbf{Cov}}_{h_{\star}}\left(\bt\right)\bSigma_{\star}^{-1}\\
\bSigma_{\star} & =\text{\textbf{Cov}}_{h_{\star}}\left(\bt\right)\label{eq: variance of hybrid =00003D variance of gaussian}
\end{align}

These last two points were already highlighted by \citet{Minka:DivMeasuresMP}.

Finally, we get a slight variant of the equations characterizing the
critical points by combining eqs. \eqref{eq: smoothed the other thing is identity}
and \eqref{eq: variance of hybrid =00003D variance of gaussian}:
\begin{equation}
\left(\boldsymbol{\text{Cov}}_{h_{\star}}\right)^{-1}\left(\bt\right)E_{h_{\star}}\left(\Ket{\bt-\bmu_{\star}}\Bra{\nabla\psi\left(\bt\right)}\right)=\bSigma_{\star}^{-1}\label{eq: the most interesting equilibrium equation}
\end{equation}

\subsection{A smoothed gradient descent}

We will now propose a smoothed gradient descent algorithm such that
all fixed-points of the algorithm will also be critical points of
$\bmu,\bS\rightarrow D_{\alpha}\left(p,q_{\bmu,\bS}\right)$.

Consider the following algorithm:
\begin{itemize}
\item Initialize the algorithm with any Gaussian approximation $q^{\left(0\right)}$
of the target distribution
\item Then loop until convergence:
\begin{enumerate}
\item Compute the current hybrid approximation of the target distribution:
\begin{equation}
h_{\alpha}\left(\bt\right)=Z_{\alpha}^{-1}p\left(\bt\right)^{1-\alpha}\left(q^{\left(t\right)}\left(\bt\right)\right)^{\alpha}
\end{equation}
\item Compute the following expected values under the current hybrid approximation
$h_{\alpha}\left(\bt\right)$:
\begin{align}
\bmu & =E_{h_{\alpha}}\left[\bt\right]\\
E\nabla & =E_{h_{\alpha}}\left[\nabla\psi\left(\bt\right)\right]\\
E\boldsymbol{H} & =\left(\text{Cov}_{h_{\alpha}}\right)^{-1}E_{h_{\alpha}}\left[\Ket{\bt-\mu}\Bra{\nabla\psi\left(\bt\right)}\right]
\end{align}
\item Compute the new Gaussian approximation using the following formula:
\begin{equation}
q^{\left(t+1\right)}=\frac{1}{Z_{q}}\exp\left(-\Bra{E\nabla}\Ket{\bt-\bmu}-\Bra{\bt-\bmu}\frac{E\boldsymbol{H}}{2}\Ket{\bt-\bmu}\right)\label{eq: the GVB updating equation-1}
\end{equation}
where $Z_{q}$ is a normalizing constant.
\end{enumerate}
\end{itemize}
Any fixed-point of this iteration must obey the following equalities:
\begin{align}
E_{h_{\star}}\left[\nabla\psi\left(\bt\right)\right] & =0\\
\left(\text{\textbf{Cov}}_{h_{\star}}\right)^{-1}E_{h_{\alpha}}\left[\Ket{\bt-\mu}\Bra{\nabla\psi\left(\bt\right)}\right] & =\left(\text{\textbf{Cov}}_{q_{\star}}\right)^{-1}
\end{align}
which is identical to the equations obeyed by critical points of $\bmu,\bS\rightarrow D_{\alpha}$
(eqs. \eqref{eq: smoothed gradient under the hybrid is 0} and \eqref{eq: the most interesting equilibrium equation}).

\section{Expectation Propagation}

Finally, we come to Expectation Propagation (EP).

In order to be able to apply EP, we have to further specify a factorization
of the target distribution:
\begin{equation}
p\left(\bt\right)=\prod_{i=1}^{n}f_{i}\left(\bt\right)
\end{equation}
Given this factorization, we can then compute ``the EP approximation
of the target distribution of $p\left(\bt\right)$ factorized as $p=\prod f_{i}$''
or, for short, the EP approximation of $p$.

We will note $\phi_{i}\left(\bt\right)=-\log\left(f_{i}\left(\bt\right)\right)$.

\subsection{The EP iteration}

All algorithms we have presented so far seek a global Gaussian approximation
of the target distribution. EP differs from this by seeking to find
instead $n$ local Gaussian approximations $q_{i}$ to approximate
each factor: $q_{i}\approx f_{i}$.

These approximations are improved iteratively according to:
\begin{itemize}
\item Initialize the algorithm with any $n$ local Gaussian approximations
$q_{i}^{\left(0\right)}\left(\bt\right)$
\item Then loop until convergence:
\begin{enumerate}
\item Select a subset $\mathcal{I}\subset\left[1,n\right]$ of indices\footnote{Several choices are possible for the selection of the subset. The
two most frequent are selecting a single index $\mathcal{I}=\left\{ i_{0}\right\} $
(corresponding to a sequential variant of EP, as originally proposed
in \citet{Minka:EP}) or the full range $\mathcal{I}=\left[1,n\right]$
(corresponding to the more modern parallel variant of EP). Minibatch
or asynchronous variants of EP are also possible. Of course, the sequence
of subsets should be such that each factor-approximation $q_{i}$
is updated regularly.}.
\item For all $i\in\mathcal{I}$ in parallel:
\begin{enumerate}
\item Compute the hybrid distribution:
\begin{equation}
h_{i}\left(\bt\right)=Z_{h_{i}}^{-1}f_{i}\left(\bt\right)\prod_{j\neq i}q_{j}^{\left(t\right)}\left(\bt\right)
\end{equation}
\item Compute the mean and covariance of the hybrid: $\left(\bmu_{h_{i}},\text{\textbf{Cov}}_{h_{i}}\right)$
\item Compute a Gaussian distribution with that mean and variance. This
Gaussian is the moment-matched Gaussian approximation of the hybrid:
\begin{equation}
q_{h_{i}}\left(\bt\right)=Z_{q_{h_{i}}}^{-1}\exp\left(-\frac{1}{2}\Bra{\bt-\bmu_{h_{i}}}\text{\textbf{Cov}}_{h_{i}}\Ket{\bt-\bmu_{h_{i}}}\right)
\end{equation}
\item Compute the new local Gaussian approximation of $f_{i}\left(\bt\right)$
given by:
\begin{equation}
q_{i}^{\left(t+1\right)}\left(\bt\right)\propto\frac{q_{h_{i}}\left(\bt\right)}{\prod_{j\neq i}q_{j}^{\left(t\right)}\left(\bt\right)}\label{eq: the value of the new local approximation}
\end{equation}
\end{enumerate}
\end{enumerate}
\end{itemize}
For a more extensive presentation of EP and the EP iteration, we refer
the interested reader to \citet{Minka:EP,Seeger:EPExpFam,BishopPRML}.

\subsection{A smoothed gradient descent}

Interestingly, the key step of the EP algorithm: the computation of
the new local approximation $q_{i}^{\left(t+1\right)}\left(\bt\right)$
from its local target $f_{i}\left(\bt\right)$ and the current values
of the other local approximations $\left(q_{j}^{\left(t\right)}\right)_{j\neq i}$,
corresponds \textbf{\uline{exactly}} to a smoothed gradient descent.
\begin{thm}
Smoothed gradient representation of the EP iteration.

In the EP iteration, the new local approximation $q_{i}^{\left(t+1\right)}\left(\bt\right)$
can also be written as:
\begin{align}
\bmu & =E_{h_{i}}\left[\bt\right]\\
E\nabla & =E_{h_{i}}\left[\nabla\phi_{i}\left(\bt\right)\right]\\
E\boldsymbol{H} & =\left(\text{\textbf{Cov}}_{h_{i}}\right)^{-1}E_{h_{i}}\left[\Ket{\bt-\mu}\Bra{\nabla\phi_{i}\left(\bt\right)}\right]\\
q_{i}^{\left(t+1\right)} & \propto\exp\left(-\Bra{E\nabla}\Ket{\bt-\bmu}-\Bra{\bt-\bmu}\frac{E\boldsymbol{H}}{2}\Ket{\bt-\bmu}\right)
\end{align}

For thoroughness, it is also important to mention that:
\begin{equation}
E\boldsymbol{H}\approx E_{h_{i}}\left(\boldsymbol{H}\phi_{i}\left(\bt\right)\right)
\end{equation}
 in the limit in which the hybrid distribution is a strongly log-concave
distribution with minimum curvature tending to $+\infty$ (\citet{dehaene2015expectation}).
\end{thm}
This theorem represents the key innovation of the present work. This
alternative formulation of the EP update might prove useful in deriving
better computational variants of the EP iteration. However, we believe
that this result is most important for theoretical investigations
of the EP algorithm. Indeed, it provides a link between EP and the
well-understood gradient descent algorithms. Furthermore, this result
also enables users of EP to have a more intuitively satisfying presentation
of how EP operates.
\begin{proof}
This result is actually fairly simple to prove. Indeed, by the definition
of $q_{h_{i}}$, $h_{i}$ and $q_{h_{i}}$ have the same mean and
variance: $\left(\bmu_{h_{i}},\text{\textbf{Cov}}_{h_{i}}\right)$.

Now, note:
\begin{align}
q_{-i}\left(\bt\right) & =\prod_{j\neq i}q_{j}^{\left(t\right)}\left(\bt\right)\\
 & \propto\exp\left(-\Bra{\boldsymbol{r}_{-i}}\Ket{\bt-\bmu}-\frac{1}{2}\Bra{\bt-\bmu}\bb_{-i}\Ket{\bt-\bmu}\right)
\end{align}

and:
\begin{equation}
q_{i}^{\left(t+1\right)}\propto\exp\left(-\Bra{\boldsymbol{r}_{i}}\Ket{\bt-\bmu}-\frac{1}{2}\Bra{\bt-\bmu}\bb_{i}\Ket{\bt-\bmu}\right)
\end{equation}

Our objective is to find a simple expression for $\boldsymbol{r}_{i}$
and $\bb_{i}$. This can be done by making use of what we call ``Stein
relationships'' (in honor of Charles Stein and his Stein's lemma,
which corresponds to the Gaussian case). Indeed, by integration by
parts, we find that for any probability distribution (with fast decrease
in the tails):
\begin{align}
E_{p}\left(\nabla\psi\left(\bt\right)\right) & =0\label{eq: the first Stein relationship}\\
E_{p}\left(\Ket{\bt-\bmu}\Bra{\nabla\psi\left(\bt\right)}\right) & =\boldsymbol{I}_{d*d}\label{eq: the second stein relationship}
\end{align}

Applying the first Stein relationship (eq. \eqref{eq: the first Stein relationship})
to the hybrid $h_{i}\propto f_{i}q_{-i}$, we get that:
\begin{align}
0 & =E_{h_{i}}\left(\nabla\left(\log h_{i}\left(\bt\right)\right)\right)\\
 & =E_{h_{i}}\left(\nabla\phi_{i}\left(\bt\right)+\left(\boldsymbol{r}_{-i}+\bb_{-i}\left(\bt-\bmu\right)\right)\right)\\
 & =E_{h_{i}}\left(\nabla\phi_{i}\left(\bt\right)\right)+\boldsymbol{r}_{-i}+\bb_{-i}\left(E_{h_{i}}\left(\bt\right)-\bmu\right)\\
 & =E_{h_{i}}\left(\nabla\phi_{i}\left(\bt\right)\right)+\boldsymbol{r}_{-i}+0\label{eq: combine me eq 81}
\end{align}

Now, we apply the first Stein relationship (eq. \eqref{eq: the first Stein relationship})
to the moment-matched approximation of $h_{i}$: $q_{h_{i}}\propto q_{i}q_{-i}$.
We get:
\begin{align}
0 & =E_{q_{h_{i}}}\left(\nabla\left(\log q_{h_{i}}\left(\bt\right)\right)\right)\\
 & =E_{q_{h_{i}}}\left(\left(\boldsymbol{r}_{i}+\bb_{i}\left(\bt-\bmu\right)\right)+\left(\boldsymbol{r}_{-i}+\bb_{-i}\left(\bt-\bmu\right)\right)\right)\\
 & =\boldsymbol{r}_{i}+\boldsymbol{r}_{-i}\label{eq: combine me eq 84}
\end{align}

By combining eqs. \eqref{eq: combine me eq 81} and \eqref{eq: combine me eq 84},
we get:
\begin{equation}
\boldsymbol{r}_{i}=E_{h_{i}}\left(\nabla\phi_{i}\left(\bt\right)\right)
\end{equation}
QED.

The proof for $\bb_{i}=\left(\text{\textbf{Cov}}_{h_{i}}\right)^{-1}E_{h_{i}}\left(\Ket{\bt-\bmu}\Bra{\nabla\phi_{i}\left(\bt\right)}\right)$
proceeds the exact same way, but by using the second Stein relationship
(eq. \eqref{eq: the second stein relationship}) to $h_{i}$ and $q_{h_{i}}$
yielding:
\begin{align}
\boldsymbol{I}_{d*d} & =E_{h_{i}}\left(\Ket{\bt-\bmu}\Bra{\nabla\phi_{i}\left(\bt\right)}\right)+\text{\textbf{Cov}}_{h_{i}}\bb_{-i}\\
 & =\text{\textbf{Cov}}_{h_{i}}\bb_{i}+\text{\textbf{Cov}}_{h_{i}}\bb_{-i}
\end{align}

Combining these equations yields the claimed result, concluding the
proof.
\end{proof}

\section{Why smoothed gradient descent leads to good approximations}

The Stein relationships we have just used in the preceeding proof
(eqs. \ref{eq: the first Stein relationship} and \ref{eq: the second stein relationship})
also provide an intuition as to why the EP and VB Gaussian approximations
(as well as the $D_{\alpha}$ minima) provide good approximations
of the target distribution. Indeed, these Stein relationships read:
\begin{align}
E_{p}\left(\nabla\psi\left(\bt\right)\right) & =0\\
E_{p}\left(\Ket{\bt-E_{p}\left(\bt\right)}\Bra{\nabla\psi\left(\bt\right)}\right) & =\boldsymbol{I}_{d*d}
\end{align}

A VB Gaussian approximation $q_{\text{VB}}$ (i.e: a minimum of $KL\left(p,q\right)$)
obeys very similar relationships:
\begin{align}
E_{q_{VB}}\left(\nabla\psi\left(\bt\right)\right) & =0\\
E_{q_{VB}}\left(\Ket{\bt-E_{p}\left(\bt\right)}\Bra{\nabla\psi\left(\bt\right)}\right) & =\boldsymbol{I}_{d*d}
\end{align}

Thus, $q_{\text{VB}}$ and $p$ have the same expected value for the
functions: $\nabla\psi\left(\bt\right)$ and $\Ket{\bt-E_{p}\left(\bt\right)}\Bra{\nabla\psi\left(\bt\right)}$
(which corresponds to $n\left(n+1\right)$ scalar equalities).

Meanwhile, a EP fixed-point with hybrids $h_{i}^{\star}$ obeys:
\begin{align}
\sum_{i}E_{h_{i}^{\star}}\left(\nabla\phi_{i}\left(\bt\right)\right) & =0\\
\sum_{i}E_{h_{i}^{\star}}\left(\Ket{\bt-E_{p}\left(\bt\right)}\Bra{\nabla\phi_{i}\left(\bt\right)}\right) & =\boldsymbol{I}_{d*d}
\end{align}
which means that the density defined by $n^{-1}\sum h_{i}^{\star}$
(where $n$ is the number of factors/factor-approximations/hybrids)
has the same expected value for the two functions we are considering.

Using this equality (or almost equality) of these expected values,
we were able last year to show that, in the classical Bayesian large-data
limit, EP and VB Gaussian approximations of a posterior distribution
are:
\begin{itemize}
\item asymptotically valid: EP and VB both correctly estimate the mean and
variance of the target distribution
\item better than the alternative Laplace approximation: the error of the
estimate of the mean is an order of magnitude better for EP and VB
than for Laplace
\end{itemize}
This result was, however, derived under restrictive assumption and
needs to be improved (\citet*{DehaeneBarthelmeNips2015}).
\end{document}